\documentclass{article}

\usepackage{microtype}
\usepackage{graphicx}
\usepackage{subfigure}
\usepackage{booktabs} % for professional tables
\usepackage{multirow}

\usepackage{hyperref}

\usepackage[accepted]{icml2025}

\usepackage{amsmath}
\usepackage{amssymb}
\usepackage{mathtools}
\usepackage{amsthm}
\usepackage{makecell}

\usepackage[capitalize,noabbrev]{cleveref}

\theoremstyle{plain}

\theoremstyle{definition}

\theoremstyle{remark}

\usepackage[textsize=tiny]{todonotes}
\definecolor{demphcolor}{RGB}{90,90,90}
\newcommand{\demph}[1]{\textcolor{demphcolor}{#1}}

\icmltitlerunning{Diffusion-Sharpening: Fine-tuning Diffusion Models with Denoising Trajectory Sharpening}

\begin{document}

\twocolumn[
\icmltitle{Diffusion-Sharpening:\\Fine-tuning Diffusion Models with Denoising Trajectory Sharpening}



\icmlsetsymbol{equal}{*}

\begin{icmlauthorlist}
\icmlauthor{Ye Tian$^{\ 1}$}{equal}
\icmlauthor{Ling Yang$^{\ 2}$}{equal}
\icmlauthor{Xinchen Zhang$^{\ 3}$}{}
\icmlauthor{Yunhai Tong$^{\ 1}$}{}
\icmlauthor{Mengdi Wang$^{\ 2}$}{}
\icmlauthor{Bin Cui$^{\ 1}$}{}
\\
$^{1}$Peking University\quad$^{2}$Princeton University\quad$^{3}$Tsinghua University
% \icmlauthor{Firstname8 Lastname8}{yyy,comp}\\
\\
Code: \href{https://github.com/Gen-Verse/Diffusion-Sharpening}{https://github.com/Gen-Verse/Diffusion-Sharpening}
\end{icmlauthorlist}

% \icmlaffiliation{yyy}{Peking University, China}
% \icmlaffiliation{comp}{Princeton University, USA}
% \icmlaffiliation{qh}{Tsinghua Univeristy, China}

\icmlcorrespondingauthor{Ling Yang}{{yangling0818@163.com}}

\icmlkeywords{Machine Learning, ICML}

\vskip 0.3in
]



\printAffiliationsAndNotice{\icmlEqualContribution} % otherwise use the standard text.

\begin{abstract}

We propose \textbf{Diffusion-Sharpening}, a fine-tuning approach that enhances downstream alignment by optimizing sampling trajectories. Existing RL-based fine-tuning methods focus on single training timesteps and neglect trajectory-level alignment, while recent sampling trajectory optimization methods incur significant inference NFE costs. Diffusion-Sharpening overcomes this by using a path integral framework to select optimal trajectories during training, leveraging reward feedback, and amortizing inference costs. Our method demonstrates superior training efficiency with faster convergence,  and best inference efficiency without requiring additional NFEs. Extensive experiments show that Diffusion-Sharpening outperforms RL-based fine-tuning methods (e.g., Diffusion-DPO) and sampling trajectory optimization methods (e.g., Inference Scaling) across diverse metrics including text alignment, compositional capabilities, and human preferences, offering a scalable and efficient solution for future diffusion model fine-tuning.
\end{abstract}

\section{Introduction}
\label{intro}

Diffusion models have emerged as a cornerstone of modern generative modeling, achieving state-of-the-art performance in tasks such as text-to-image synthesis and video generation ~\citep{ho2020denoising,song2020score,sohl2015deep, ramesh2022hierarchical, rombach2022high, ho2022imagen, blattmann2023align, zhang2024realcompo}. Despite their success, fine-tuning these models to align with diverse and nuanced user preferences remains a fundamental challenge, particularly in domains requiring fine-grained or domain-specific control over generated outputs.  

\begin{figure}[ht]
    \centering
    \includegraphics[width=\linewidth]{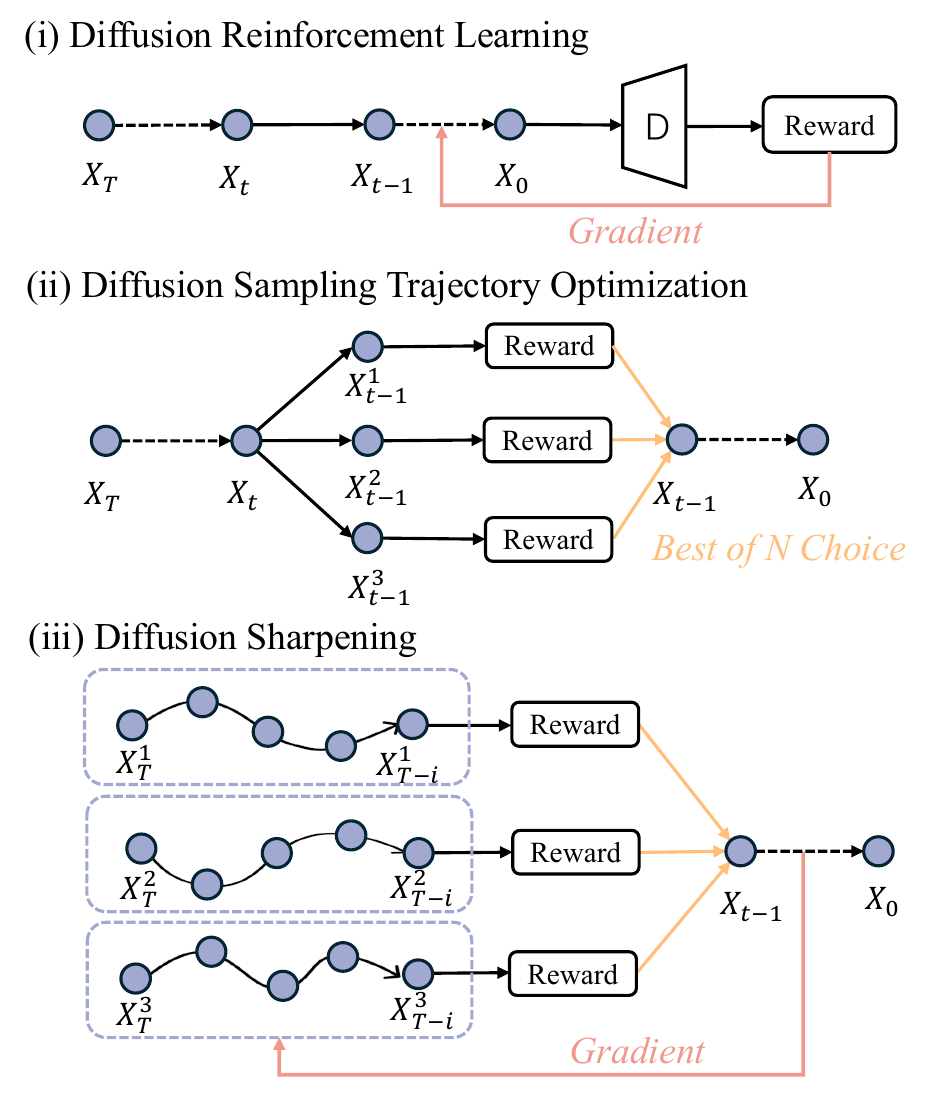}
    % \vspace{-2em}
    \caption{Comparison of Three Diffusion-Based Methods for Reward-Driven Optimization: (i) Diffusion Reinforcement Learning, (ii) Diffusion Sampling Trajectory Optimization, and (iii) Diffusion Sharpening.}
    \label{fig:intro}
    \vspace{-2em}
\end{figure}

% A promising direction for aligning diffusion models with user preferences involves fine-tuning them using reinforcement learning (RL) to optimize reward signals that capture user-defined objectives~\citep{black2024training, wallace2024diffusion}. However, these methods are hindered by their reliance on large-scale human preference datasets, which are expensive and labor-intensive to construct. Moreover, many datasets provide only a single image per prompt, making it impractical to generate the preference pairs necessary for RL-based fine-tuning.  

Fine-tuning diffusion models to align with predefined evaluation criteria or human preferences remains a key challenge. A promising approach involves fine-tuning these models using reinforcement learning (RL) through gradient-based optimization during training to optimize reward signals that reflect user-defined objectives~\citep{black2024training, wallace2024diffusion, prabhudesai2023aligning, xu2024imagereward,zhang2024itercomp}, as shown in \cref{fig:intro} (i).While effective with large-scale curated datasets, these methods focus on optimizing a \textbf{single timestep}'s output and overlook the potential for optimizing the entire sampling trajectory.

Recent approaches extend optimization to the backward denoising process, enabling real-time adjustments during diffusion sampling and performing progressive trajectory refinement. As illustrated in \cref{fig:intro} (ii), these sampling trajectory optimization methods~\citep{kim2024free, yeh2024training, ma2025inference} demonstrate that intermediate states along the trajectory can guide generative improvements. However, these methods incur significant computational overhead, with high-quality generation taking up to 40 minutes per image~\citep{yeh2024training}, making them impractical for real-world use.

To address these limitations, we propose \textbf{Diffusion-Sharpening}, a fine-tuning framework that enhances diffusion model alignment by optimizing the sampling trajectory, as shown in \cref{fig:intro}(iii). During training, we sample multiple trajectories and compute rewards through path integration, guiding the model to optimize towards the best trajectory. We introduce two implementations: (1) \textbf{SFT-Diffusion-Sharpening}, which uses a pre-existing image-text dataset for supervised fine-tuning, enabling optimization with any reward model; and (2) \textbf{RLHF-Diffusion-Sharpening}, which uses online methods to generate positive and negative samples from denoising outputs, achieving self-guided learning and improved alignment with any reward model through DPO loss.

We experimentally demonstrate the effectiveness of our method, showing its efficient convergence during training compared to standard fine-tuning, as well as its high efficiency during inference without the need for additional search costs. Furthermore, Diffusion-Sharpening consistently outperforms RL-based fine-tuning methods and sampling trajectory optimization methods across a range of image generation metrics, including text alignment, compositional abilities, and human preferences.

Our contributions are summarized as follows:
\begin{itemize}
    \item We introduce \textbf{Diffusion-Sharpening}, a fundamental and effective trajectory-level optimization-based fine-tuning method that aligns diffusion models with arbitrary pre-defined rewards.
    \item We develop \textbf{SFT-Diffusion-Sharpening} and \textbf{RLHF-Diffusion-Sharpening}, with the former providing a more efficient SFT pipeline, while the latter eliminates the need for dataset curation in DPO training.
    \item Compared to previous fine-tuning-based and sampling trajectory optimization methods, our approach achieves the best training and inference efficiency, while setting state-of-the-art performance across diverse metrics, including text alignment, compositional capabilities, and human preferences. 
\end{itemize}

\section{Related Work}

\subsection{Diffusion Alignment}
Diffusion alignment aims to align model outputs with user preferences by integrating reinforcement learning (RL) into diffusion models to enhance generative controllability \citep{wallace2024diffusion,xu2024imagereward,zhang2025itercomp,ueharafeedback,yang2024using}. DDPO~\citep{black2024training} uses predefined reward functions to fine-tune diffusion models for specific tasks, such as compressibility. In contrast, DPOK~\citep{fan2024reinforcement} utilizes feedback from AI models trained on large-scale human preference datasets. 
% Similarly, SEIKO~\citep{ueharafeedback} constructs custom reward models based on extensive human feedback data to guide the generative process.
An alternative to predefined rewards is direct preference optimization (DPO). 
Diffusion-DPO~\citep{wallace2024diffusion} extends DPO~\citep{clark2024directly} to diffusion models by directly utilizing preference data for fine-tuning, thereby eliminating the need for predefined reward functions. Despite its potential, Diffusion-DPO relies on large-scale preference datasets and still fails to handle complex generation scenarios.
Recent IterComp \citep{zhang2024itercomp} address these challenges by gathering composition-aware preference data from a set of open-sourced models and aligning with the collected preferences iteratively.
% In contrast, D3PO~\citep{yang2024using} replaces offline datasets with online human feedback collected dynamically during training. 

\subsection{Diffusion Trajectory Forward Optimization}
Forward optimization in diffusion trajectories focuses on refining the forward process through carefully designed transition kernels or data-dependent initialization distributions~\citep{liu2022flow,hoogeboom2022blurring,dockhorn2021score,lee2021priorgrad,karras2022elucidating,yang2024cross}. For instance, Rectified Flow~\citep{liu2022flow} and Consistency Flow Matching \citep{yang2024consistency} learns a straight path connecting the data distribution and the prior distribution, effectively simplifying the denoising process. Grad-TTS~\citep{popov2021grad} and PriorGrad~\citep{lee2021priorgrad} introduce conditional forward processes with data-dependent priors, specifically designed for audio diffusion models. Other methods like ContextDiff \citep{yang2024cross} focus on parameterizing the forward process with additional neural networks. For example, Diffusion Models for Video Generation~\citep{zhang2021diffusion}, Maximum Likelihood Training for Score-based Diffusion Models~\citep{kim2022maximum}, and Variational Diffusion Models (VDM)~\citep{kingma2021variational} employ neural architectures to enhance the forward trajectory. 

\subsection{Diffusion Trajectory Sampling Optimization}
Beyond forward optimization, recent research has explored real-time optimization during the sampling process, incorporating stochastic optimization techniques to guide the backward sampling trajectory. For instance, MBD~\citep{pan2024modelbased} utilizes score functions to direct the sampling path in the backward process. Similarly, in music generation tasks, SCG~\citep{huang2024symbolic} employs stochastic optimization to leverage non-differentiable reward functions. Demon~\citep{yeh2024training} focuses on optimizing the sampling process to concentrate sampling density in regions with high rewards during inference. Free$^{2}$Guide~\citep{kim2024free} uses path integral control to provide gradient-free, non-differentiable reward guidance, enabling the alignment of generated videos with textual prompts without requiring additional model training. Inference-Scaling~\citep{ma2025inference} employs a verifier and search algorithm to scale diffusion inference beyond NFEs.  

While these approaches demonstrate significant potential, they often incur substantial computational overhead due to the extra steps required for calculating intermediate rewards during inference. For example, Demon~\citep{yeh2024training} and Inference-Scaling~\citep{ma2025inference} may require up to 1000x the inference cost per image to achieve optimal performance. This significant increase in computational cost considerably slows down the generation process, limiting their practicality for real-world applications.

\section{Method}
\begin{figure*}
    \centering
    \includegraphics[width=0.9\linewidth]{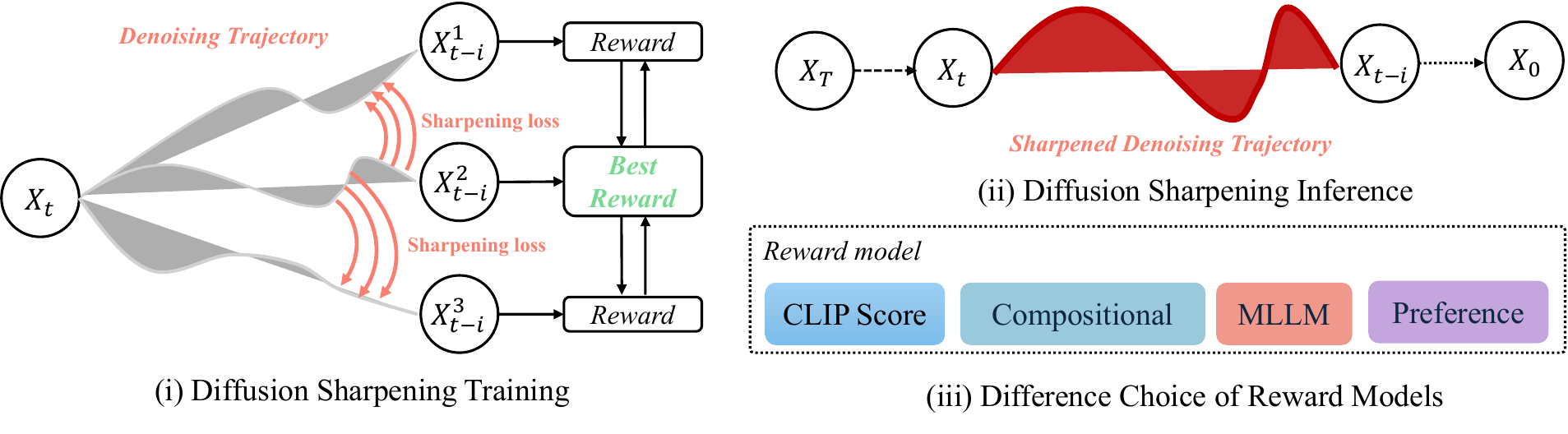}
    \caption{Overview of Our Diffusion Sharpening Framework: (i) Training, (ii) Inference, and (iii) Reward Model Selection}
    \label{fig:method}
    \vspace{-1.5em}
\end{figure*}

\subsection{Preliminaries}
Diffusion Probabilistic Models ~\citep{sohl2015deep, song2020score, ho2020denoising} learns a stochastic process by iteratively denoising random noise generated by the forward diffusion process. Specifically, for any $t \in (0, T]$, the transition distribution is defined as:
\begin{align}\label{eq:tran}
p(\mathbf x_t|\mathbf x_0, c) = p(\mathbf x_t|\mathbf x_0) = \mathcal N(\alpha_t \mathbf x_0, \sigma_t^2 \mathbf I),
\end{align}
where $\mathbf x_0 \in \mathbb R^D$ is a $D$-dimensional data signal variable with an unknown distribution $p_0(\mathbf x_0|c)$, $c \sim q(c)$ is the given condition, and $\alpha_t, \sigma_t \in \mathbb R^+$ are noise scheduler.

Foundational works~\citep{kingma2021variational, song2020score} have analyzed the underlying stochastic differential equation (SDE) and ordinary differential equation (ODE) formulations for DPM. The forward and reverse dynamics are given for any $t \in [0, T]$ as:
\begin{align}
\mathrm d \mathbf x_t &= f( \mathbf x_t) \mathrm d t + g(t) \mathrm d \mathbf w_t, \quad \mathbf x_0 \sim p_0(\mathbf x_0 | c), \label{eq:forward} \\
\mathrm d \mathbf x_t &= \big[f(\mathbf x_t) - g^2(t) \nabla_{\mathbf x_t} \log p_t(\mathbf x_t | c)\big] \mathrm d t + g(t) \mathrm d \bar{\mathbf w}_t, \label{eq:reverse}
\end{align}
where $\mathbf w_t$ and $\bar{\mathbf w}_t$ are standard Wiener processes in forward and reverse time, respectively, and $f$ and $g$ are functions defined in terms of $\alpha_t$ and $\sigma_t$. 

Practically, DPM performs sampling by solving either the reverse SDE or ODE backward from $T$ to $0$.  To facilitate this, a neural network $\boldsymbol \epsilon_\theta(\mathbf x_t, c, t)$, known as the noise prediction model, is introduced to approximate the conditional score function based on $\mathbf x_t$ and $c$ at time $t$. 
Specifically, $\boldsymbol \epsilon_\theta(\mathbf x_t, c, t) = -\sigma_t \nabla_{\mathbf x_t} \log p_t(\mathbf x_t | c)$, and its parameters $\theta$ are optimized via the objective:
\begin{align}\label{eq:dpm_loss}
\mathbb E_{\mathbf x_0, \boldsymbol \epsilon, c, t} \left[\omega_t \|\boldsymbol \epsilon_\theta(\mathbf x_t, c, t) - \boldsymbol \epsilon\|_2^2 \right],
\end{align}
where $\omega_t$ is a weighting function, $\boldsymbol \epsilon \sim \mathcal N(\mathbf 0, \mathbf I)$, $c \sim q(c)$, $\mathbf x_t = \alpha_t \mathbf x_0 + \sigma_t \boldsymbol \epsilon$, and $t \sim \mathcal U[0, T]$.

\subsection{Diffusion Sharpening}
\label{method}

In autoregressive language models, performance can be improved through "self-improvement," where the model itself acts as a validator. Specifically, a base model \(\pi_{\text{base}}\colon X \to \Delta(Y)\), representing a conditional distribution, evaluates generated sequences. We refer to \textbf{sharpening} as training the model to produce outputs with higher conditional probabilities shifts the model's distribution towards more confident and higher-quality responses. Formally, a sharpening model $\hat{\pi}(x)$ is one that (approximately) maximizes the self-reward toward responses that maximize a self-reward \(r_{\text{self}}\):  
\[
\hat{\pi}(x) \approx \arg\max_{y \in \mathcal{Y}} r_{\text{self}}(y \mid x; \pi_{\text{base}}).
\]  
While sharpening in language models focuses on sequence-level optimization, diffusion alignment typically fine-tunes individual trajectory points, which may lead to suboptimal results. The lack of trajectory-level feedback exposes the generative process to stochastic noise and inconsistencies along the sampling path.  

To address these challenges, we propose Diffusion-Sharpening, which leverages online alignment techniques for fine-tuning diffusion models. First, we approximate \(x_0\) from intermediate states to evaluate the reward for \(x_t\). Then, we perform reward evaluation along the sampling trajectory. To implement this, we introduce two fine-tuning strategies: \textit{SFT-Diffusion-Sharpening} and \textit{RLHF-Diffusion-Sharpening}.
\paragraph{Approximate \(x_0\) for Reward Evaluation}

We leverage techniques from EDM~\citep{karras2022elucidating} to approximate the posterior distribution of \(x_0\) given an intermediate state \(x_t\). Starting from the reverse-time SDE\cref{eq:reverse}, we have
\begin{align} \mathbf{x}_0 = \mathbf{x}_t + \int_t^0 \mathbf{f}(\mathbf{x}|c)\mathrm{d}u +  g(u)\mathrm{d}\bar{\mathbf w}_u, \label{eq:x0_approx} \end{align}

where $\mathbf{f}(\mathbf{x}|c) = f(\mathbf x_t) - g^2(t) \nabla_{\mathbf x_t} \log p_t(\mathbf x_t | c)$ represents the drift term , while \(\bar{\mathbf w}_u\) denotes the stochastic noise component. To simplify this estimation, as shown in~\citep{song2020score, karras2022elucidating}, the reversed-time SDE reduces to PF-ODE when $\beta \equiv 0$. For each $t$, a diffeomorphic relationship exists between a noisy sample $\mathbf{x}_t$ and a clean sample $\mathbf{x}_0$ generated by PF-ODE~\citep{yeh2024training}:
\begin{align}
\label{reward_signle}
    \mathbf{c}(\mathbf{x}_t, t) := \mathbf{x}_0 = \mathbf{x}_t + \int_t^0( -u \nabla_{\mathbf{x}_u} \log p(\mathbf{x}_u|c)) \, \mathrm{d}u.
\end{align}

For any timestep \(x_t\) in the diffusion process and a given condition \(c\), the reward \(R(x_t, c)\) is defined as:  
\begin{align}
\label{reward}
R(x_t, c) = \text{Reward}(\mathbf{c}(x_t, t)),
\end{align}

where \(\text{Reward}(\cdot)\) represents any reward model, which can be implemented using various forms, such as a differentiable neural network, a human feedback-based scoring function, or even a non-differentiable external model like a multimodal LLM.  

\paragraph{Trajectory-Level Reward Aggregation}

Fine-tuning based on a single trajectory point is often insufficient, as it is highly sensitive to stochastic perturbations in the noise distribution. To address this limitation, we computed and aggregated rewards over selected diffusion sampling trajectories $\tau$ rather than individual $x_t$. Specifically, this involves evaluating the reward for different sampled trajectories and selecting the optimal one based on cumulative feedback:  
\begin{align}
    \label{eq:traj}
    \hat{\tau} = \arg\max_{\tau \in \mathcal{T}} \sum_{t \in \tau} R(x_t, c),
\end{align}
where \(\mathcal{T}\) denotes the set of possible trajectories.  

This approach ensures that the diffusion model learns to generate sampling paths with consistently high rewards, leading to improved sample quality and more robust generative behavior. 

\subsection{Algorithms for Diffusion Sharpening}
In this section, we present two families of self-improvement algorithms for diffusion sharpening: \textit{SFT Diffusion Sharpening}, which filters high-reward responses and performs \textbf{online} fine-tuning using standard supervised learning pipelines, and \textit{RLHF Diffusion Sharpening}, which refines the sampling trajectory by \textbf{online} optimizing winning and losing sets through reinforcement learning techniques, such as Diffusion-DPO~\citep{wallace2024diffusion}.

\paragraph{SFT Diffusion Sharpening}
In the language model framework, SFT-Sharpening~\citep{huang2024self} filters responses with large self-reward values and applies standard supervised fine-tuning to the resulting high-quality samples. Similarly, in the pretraining or supervised fine-tuning (SFT) of text-to-image diffusion models, a large image-text dataset is filtered through selected reward models, retaining the highest-scoring image-text pairs for training. However, this direct fine-tuning process only captures the preferences of the final output generated by the diffusion model, relying solely on a single random timestep and backpropagating with \(v\)-loss~\citep{salimans2022progressive} or \(\epsilon\)-loss~\citep{ho2020denoising}.  
We argue that this approach fails to fully exploit the potential of each sample (image, text), as one timestep's $v$-prediction or $\epsilon$-prediction  cannot represent the entire denoising trajectory.  

To address this limitation, as discussed in \cref{method}, we propose a redesigned \textit{SFT Diffusion Sharpening} process that fully utilizes the sampling trajectory for each sample (image, text), prestented in \cref{alg:diffusion_sft_sharpening}. Specifically, consider a collection of image-text pairs \((x, c)\) from a dataset \(D\). For each prompt \((x, c)\), we randomly sample \(n\) noise vectors \(z_1, \dots, z_n \sim \mathcal{N}(0, 1)\), and then randomly select a timestep \({t}\). Noise is added to the image \(x\) to generate \(x^i_t\), where \(i \in \{1, \dots, N\}\). We then perform sampling on the noisy images \(x^i_t\) for \(m\) steps and collect the corresponding sampling trajectories. Afterward, we select the optimal trajectory based on reward feedback. Finally, we backpropagate the gradients using the loss from the \(m\)-step path
\begin{align}
\label{sft_loss}
    L = \mathbb{E}_{\mathbf{x}_T, \boldsymbol{\epsilon}, c, T \in [t, t - m]} \left[ \omega_T \| \boldsymbol{\epsilon}_\theta(x_T, c, T) - \boldsymbol{\epsilon} \|_2^2 \right]\,.
\end{align}

\begin{algorithm}[htb]
   \caption{SFT Diffusion Sharpening}
   \label{alg:diffusion_sft_sharpening}
\begin{algorithmic}
   \STATE {\bfseries Input:} dataset \(D\), number of samples \(n\), number of steps \(m\), reward model \(R\), diffusion model $M_\theta$, learning rate $\eta$
   \FOR{each image-text pair \((x, c)\) in \(D\)}
       \STATE Sample \(n\) random noise vectors \(z_1, \dots, z_n \sim \mathcal{N}(0, 1)\)
       \STATE Randomly select a timestep \(t\)
       \STATE Add noise to the image \(x\) to generate \(x^i_t\) for \(i \in \{1, \dots, N\}\)
       \FOR{each noisy image \(x^i_t\)}
           \STATE Perform \(m\) steps of sampling from \(x^i_t\)
           \STATE Calculate $R(x_t, c)$ in \cref{reward}
           \STATE Collect the sampling trajectory \(\tau_i = \{x_{t_k}\}_{k=1}^m\)
       \ENDFOR
       \STATE Select the optimal trajectory \(\hat{\tau}\) with \cref{eq:traj}
       \STATE Compute the loss $L$ in \cref{sft_loss} 
       \STATE $M_\theta \leftarrow M_\theta - \eta \nabla_\theta L$
   \ENDFOR

\end{algorithmic}

\end{algorithm}

\paragraph{RLHF Diffusion Sharpening}
RLHF Diffusion Sharpening~\cref{alg:rlhf_diffusion_sharpening} aims to optimize a conditional distribution $p_\theta(x_t|c)$ such that the reward model $R(x_t, c)$ defined on it is maximized while regularizing the KL-divergence from a reference distribution $p_{ref}$

\begin{align}
\max_{\theta} \mathbb{E}_{c \sim \mathcal{D}_c, x_0 \sim p_{\theta}(x_0 | c)} & \left[ r(c, x_0) \right]  \notag \\
- \beta & \mathbb{D}_{KL} \left[ p_{\theta}(x_0 | c) \parallel p_{\text{ref}}(x_0 | c) \right] 
\end{align}

For its efficiency, we adopt Diffusion-DPO~\citep{wallace2024diffusion} to implement RLHF Diffusion Sharpening, aiming to fully leverage the model’s self-evolution capabilities. Instead of relying on predefined image-text pairs, we construct the dataset online by generating latent samples and applying noise perturbations during training. Similar to SFT Diffusion Sharpening, after selecting a set of noisy samples \(x^i_t\) and their corresponding trajectories, we use a reward model to identify the best and worst trajectories, \(\tau_w\) and \(\tau_l\), respectively. To maximize the use of prior reward information, we optimize the model using the reward-modulated DPO loss~\citep{gao2024rebel}. 

\begin{align}
    \label{dpo-loss}
    L_{\text{RLHF}}(\theta) &=  \mathbb{E}_{x_w \in \tau_w, x_l \in \tau_l, c} \notag \\
    & \quad \left[ \log \sigma \left( \beta \log \frac{p_\theta(x_w \mid c)}{p_{\text{ref}}(x_w \mid c)} - \beta \log \frac{p_\theta(x_l \mid c)}{p_{\text{ref}}(x_l \mid c)} \right)   \right.  \notag \\
    & \quad \left. \left.  - (R(x_w, c) - R(x_l, c)) \right) \right]
\end{align}

\begin{algorithm}[htb]
   \caption{RLHF Diffusion Sharpening}
   \label{alg:rlhf_diffusion_sharpening}
\begin{algorithmic}
   \STATE {\bfseries Input:} prompt dataset \(D\), number of samples \(n\), number of steps \(m\), reward model \(R\), diffusion model \(M_\theta\), learning rate \(\eta\)
   \FOR{each training iteration}
       \STATE Sample prompt \(c\) and generate latents with \(M_\theta\)
       \STATE Sample \(n\) random noise vectors \(z_1, \dots, z_n \sim \mathcal{N}(0, 1)\)
       \STATE Randomly select a timestep \(t\) and add noise to generated latents for noisy latent samples \(x^i_t\)
       \FOR{each noisy sample \(x^i_t\)}
           \STATE Perform \(m\) steps of sampling from \(x^i_t\)
           \STATE Evaluate reward \(R(x_t, c)\) using the reward model
           \STATE Collect the sampling trajectory \(\tau_i = \{x_{t_k}\}_{k=1}^m\)
       \ENDFOR
       \STATE Identify best and worst trajectories:
       \STATE $\tau_w = \arg\max_{\tau \in \mathcal{T}} \sum_{t \in \tau} R(x_t, c),$
       \STATE $\tau_l = \arg\min_{\tau \in \mathcal{T}} \sum_{t \in \tau} R(x_t, c),$
       \STATE Compute \(L_{\text{RLHF}}(\theta)\) using \cref{dpo-loss}
       \STATE Update model parameters: \(M_\theta \leftarrow M_\theta - \eta \nabla_\theta L_{\text{DPO}}\)
   \ENDFOR
\end{algorithmic}
\end{algorithm}

\section{Experiments}

\begin{table*}[htb]
\centering
\caption{Comparison of Model Performance across Multiple Metrics}
\resizebox{\linewidth}{!}{
\begin{tabular}{lcccccccccc}
\toprule
\label{tab:main}
\multirow{2}{*}{\textbf{Model}} & \multirow{2}{*}{\textbf{CLIP Score}} & \multicolumn{6}{c}{\textbf{T2I-Compbench}} & \multirow{2}{*}{\textbf{Aesthetic}} & \multirow{2}{*}{\textbf{ImageReward}} & \multirow{2}{*}{\textbf{MLLM}} \\
\cmidrule(lr){3-8}
 &  & \textbf{Color} & \textbf{Shape} & \textbf{Texture} & \textbf{Spatial} & \textbf{Non-Spatial} & \textbf{Complex} & \\
\midrule
SDXL & 0.322 & 0.6369 & 0.5408 & 0.5637 & 0.2032 & 0.3110& 0.4091 & 5.531 & 0.780 & 0.780 \\
\hline
\multicolumn{11}{c}{{ \demph{\textit{\textbf{Fine-tuning based Methods}} }} }\\
\hline

Standard Fine-tuning & 0.325 & 0.6437 & 0.5771 & 0.5692 & 0.2084 & 0.3147 & 0.4100 & 5.556 & 0.791 & 0.784 \\
Diffusion DPO~\citep{wallace2024diffusion} & 0.334 & 0.6602 & 0.5553 & 0.5640 & 0.2112 & 0.3180 & 0.4055 & 5.754 & 1.352 & 0.864 \\
DDPO~\citep{black2024training} & 0.324 & 0.6435 & 0.5365 & 0.5531 & 0.2030 & 0.3142 & 0.4024 & 5.640 & 0.910 & 0.791 \\
D3PO~\citep{yang2024using} & 0.328 & 0.6434 & 0.5435 & 0.5657 & 0.2114 & 0.3153 & 0.4102 & 5.528 & 0.982 & 0.785 \\

\textcolor{black}{IterPO~\citep{zhang2024itercomp}} & 0.335 & 0.6637 & 0.5593 & 0.6167 & 0.2128 & 0.3207 & 0.4377 & 5.923 & 1.408 & 0.884 \\
\hline
\multicolumn{11}{c}{{ \demph{\textit{\textbf{Sampling Trajectory Optimization Methods}} }} }\\
\hline

Free$^{2}$Guide~\citep{kim2024free} & 0.325 & 0.6321 & 0.5386 & 0.5548 & 0.2050 & 0.3125 & 0.4082 & 5.560 & 0.873 & 0.786 \\
Demon~\citep{yeh2024training} & 0.325 & 0.6502 & 0.5507 & 0.5602 & 0.2150 & 0.3158 & 0.4070 & 5.630 & 1.243 & 0.300 \\
Inference Scaling~\citep{ma2025inference} & 0.328 & 0.6550 & 0.5527 & 0.5700 & 0.2204 & 0.3168 & 0.4265 & 5.752 & 1.329 & 0.872 \\

\midrule
SFT Diffusion Sharpening & 0.334 & 0.6578 & 0.5692 & 0.5733 & 0.2120 & 0.3185 & 0.4125 & 5.785 & 1.301 & 0.864 \\
\textbf{RLHF Diffusion Sharpening} & \textbf{0.338} & \textbf{0.6841} & \textbf{0.5680} & \textbf{0.6401} & \textbf{0.2134} & \textbf{0.3220} & \textbf{0.4498} & \textbf{5.956} & \textbf{1.445} & \textbf{0.921} \\
\bottomrule
\end{tabular}}
\label{tab:finetuning_comp}
\end{table*}

\subsection{Implemention Details}
\paragraph{Baseline Models}  
We conduct diffusion sharpening fine-tuning on SDXL~\citep{podell2023sdxl} for a fair comparison, using the default configuration with a DDIM Scheduler, \(T = 50\) steps, and classifier-free guidance with a scale of \(w = 5\). For comparison with fine-tuning methods, we select five established approaches: (1) Standard Fine-tuning\footnote{\url{https://github.com/huggingface/diffusers/blob/main/examples/text_to_image/train_text_to_image_sdxl.py}}, traditional fine-tuning using a predefined image-text dataset; (2) Diffusion-DPO~\citep{wallace2024diffusion}, fine-tuning based on human preference datasets; (3) DDPO~\citep{black2024training}, reward model-based reinforcement fine-tuning; (4) D3PO~\citep{yang2024using}, fine-tuning using human feedback without a reward model; \textcolor{black}{and (5) IterPO~\citep{zhang2024itercomp}, iterative alignment of composition-aware model preferences introduced in IterComp~\citep{zhang2024itercomp}.}
For comparison with sampling trajectory optimization methods, we select: (1) Demon~\citep{yeh2024training}, which recalculates the optimal noise at each denoising timestep to optimize inference; (2) Free$^{2}$Guide~\citep{kim2024free}, an inference optimization method for video generation that searches for optimal noise over 1/10 of the timesteps; and (3) Inference Scaling~\citep{ma2025inference}, which performs inference using a search and verifier mechanism. These methods are adapted to SDXL with default settings as described in their respective papers. More Details are provided in \cref{sec:baseline}.

\paragraph{Datasets}  
For SFT Diffusion-Sharpening, we use two high-quality text-to-image datasets: JourneyDB~\citep{pan2023journeydb} and Text-to-Image-2M~\citep{zk_2024}, which contain a large number of image-text pairs, ideal for evaluating the benefits of sharpening over baseline SFT methods. Additionally, we employ the domain-specific dataset Pokemon-Blip-Caption~\citep{pokeman} to assess sharpening's effectiveness in personalized scenarios, measuring its adaptability while preserving output quality. For RLHF Diffusion-Sharpening, no image data is required during training as online optimization relies solely on prompts. We randomly sample 10,000 prompts from DrawBench~\citep{saharia2022photorealistic}, DiffusionDB~\citep{wang2022diffusiondb}, and prompts from the SFT datasets for fine-tuning. More details are included in \cref{app:dataset}

\paragraph{Reward Models}  
We evaluate the performance of various reward models in diffusion sharpening, analyzing their effectiveness and efficiency across tasks: (1) CLIP Score~\citep{radford2021learning}, used to evaluate text-image alignment, (2) Compositional rewards from IterComp~\citep{zhang2025itercomp}, which assess the model's ability to handle compositional prompts such as object relationships and attributes, (3) MLLM grader~\citep{ma2025inference}, specifically prompted GPT-4o, detailed in \cref{app:mllm}, which provide holistic image scoring across multiple dimensions to improve overall quality, and (4) Human Preferences. We employ ImageReward~\citep{xu2024imagereward}, a reward model trained to align with human preferences, to evaluate satisfaction with text-image alignment, aesthetic quality, and harmlessness.

\paragraph{Evaluation Metrics}
We use several key metrics to evaluate the performance of our models: (1) CLIP Score~\citep{radford2021learning}, which measures text-image alignment, (2) Aesthetic Score from DrawBench~\citep{saharia2022photorealistic}, assessing the visual appeal and quality of the generated image (3) T2I-Compbench~\citep{huang2023t2i}, to evaluate the compositional capabilities and (4) ImageReward~\citep{xu2024imagereward}, which evaluates how well the generated images align with human preferences, including text-image consistency, aesthetic quality, and overall satisfaction. We also report scores used in the MLLM grader for overall evalution.  Additionally, due to the inherent subjectivity in evaluating image generation tasks, we conducted an extensive user study to complement our quantitative metrics in \cref{app:user}.

\subsection{Main Results}

\paragraph{Comparison with Fine-tuning based Methods}
\begin{figure*}
    \centering
    \includegraphics[width=\linewidth]{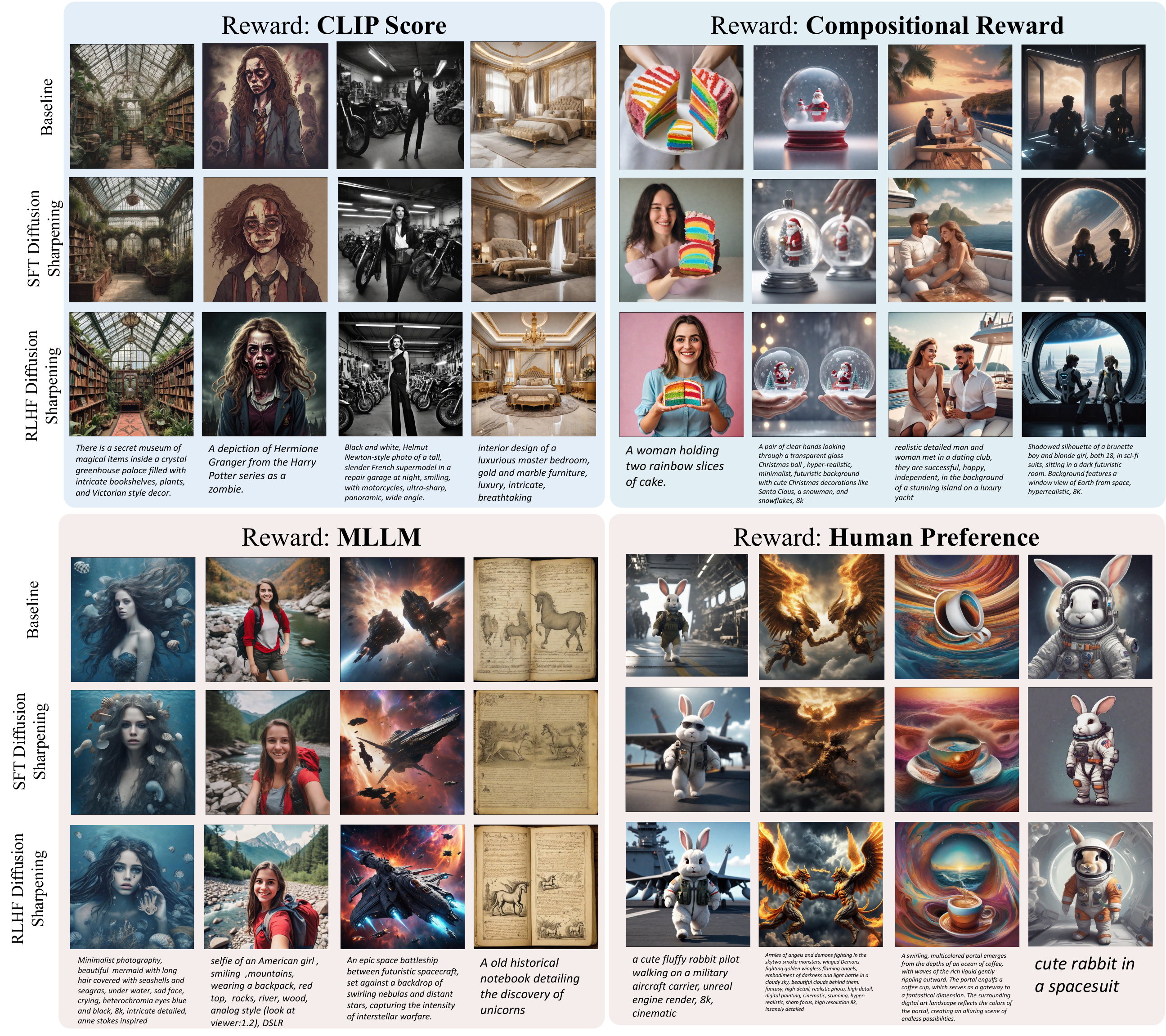}
    \caption{Qualitative results comparing Diffusion Sharpening methods using different reward models. The images show the generated results with CLIP Score, Compositional Reward, MLLM, and Human Preferences as reward models, showcasing the effectiveness of SFT Diffusion Sharpening and RLHF Diffusion Sharpening in diffusion finetuning.}
    \label{fig:demo}
\end{figure*}
\begin{figure*}
    \centering
    \includegraphics[width=0.9\linewidth]{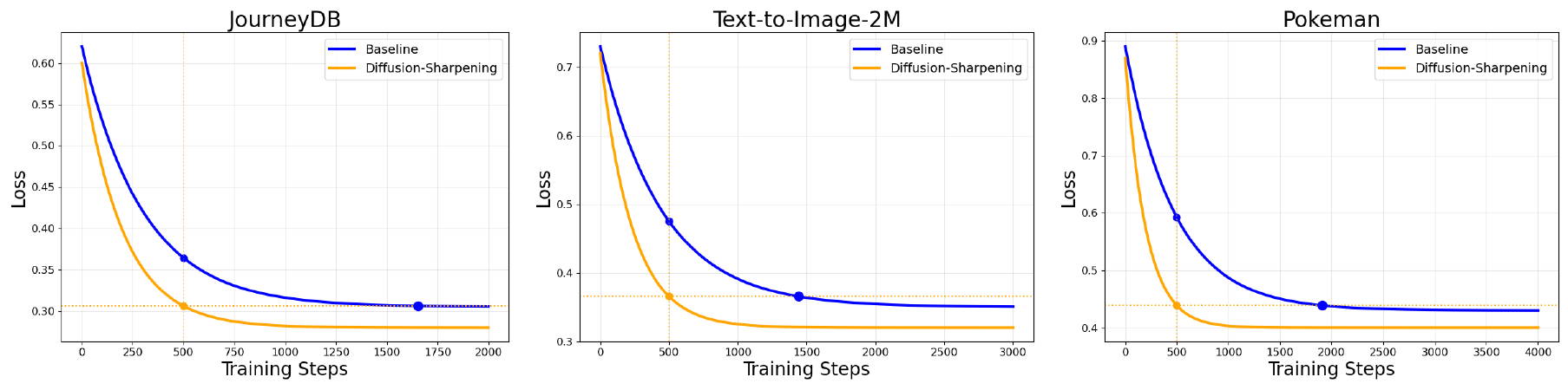}
    \caption{SDXL Finetuning Loss across Difference Datasets. Here "Diffusion-Sharpening" represents SFT Diffusion-Sharpening specifically.}
    \label{fig:loss}
\end{figure*}

In the quantitative analysis, we compare various methods, as shown in \cref{tab:main}. We train Diffusion-Sharpening on different reward models and report the corresponding evaluation metrics. Notably, the Aesthetic score is derived as the average result across 4 rewards' corresponding model. Our approach outperforms Diffusion-DPO and D3PO in human preference evaluations and generalizes to any reward model. Compared to DDPO, which also uses reward model-based fine-tuning, our sharpening method optimizes the most relevant reward path, further improving overall performance. \textcolor{black}{Compared to IterPO, our method achieves improved image compositionality, further enhancing model's alignment with complex compositional prompts.} As seen in the table, RLHF-Diffusion-Sharpening consistently achieves top results across all evaluation metrics, demonstrating exceptional generalization and adaptability to diverse reward models. Qualitative results, presented in \cref{fig:demo}, show that our model leverages multiple reward models tailored to specific needs, improving text-image alignment, compositional abilities, human preferences, and MLLM assessments. 
RLHF-Diffusion-Sharpening, in particular, excels in both qualitative and quantitative performance. These improvements stem from the base model's extensive pretraining on large datasets. In SFT-Sharpening, the standard epsilon-loss converges quickly, leaving little room for further enhancement. However, RLHF-Diffusion-Sharpening, through DPO loss, better separates good and bad trajectories, offering greater optimization potential.

\paragraph{Comparison with Sampling Trajectory Optimization Methods}  
We also compare with sampling trajectory  optimization methods. As shown in \cref{tab:main}, Free$^{2}$Guide provides slight improvements in image generation, but its performance is limited. Demon and Inference Scaling improve by increasing inference steps (NFE), but our method achieves superior quantitative results while effectively amortizing inference costs, demonstrating efficiency and validity.

% \subsection{Generalizing to Arbitrary Reward Functions}

\subsection{Model Efficiency}
\paragraph{Training Efficiency}

Diffusion Sharpening significantly enhances model efficiency through trajectory-level optimization. During the training phase, we set \(\tau = 3\) and \(n = 3\) random noise vectors, with a learning rate of \(1 \times 10^{-6}\), comparing it to the standard SDXL fine-tuning pipeline. We reported the fitted loss curve in \cref{fig:loss}. As is shown, Diffusion-Sharpening leads to faster convergence, typically within 500 to 1000 steps, whereas the baseline requires 1000 to 1500 steps to achieve similar results. The training curve for Diffusion Sharpening is smoother and achieves better final convergence with a lower final loss. These results demonstrate that Diffusion Sharpening enables faster, more stable, and superior fine-tuning compared to standard diffusion pipelines.
\paragraph{Inference Efficiency}
\begin{figure}[ht]
    \centering
    \includegraphics[width=\linewidth]{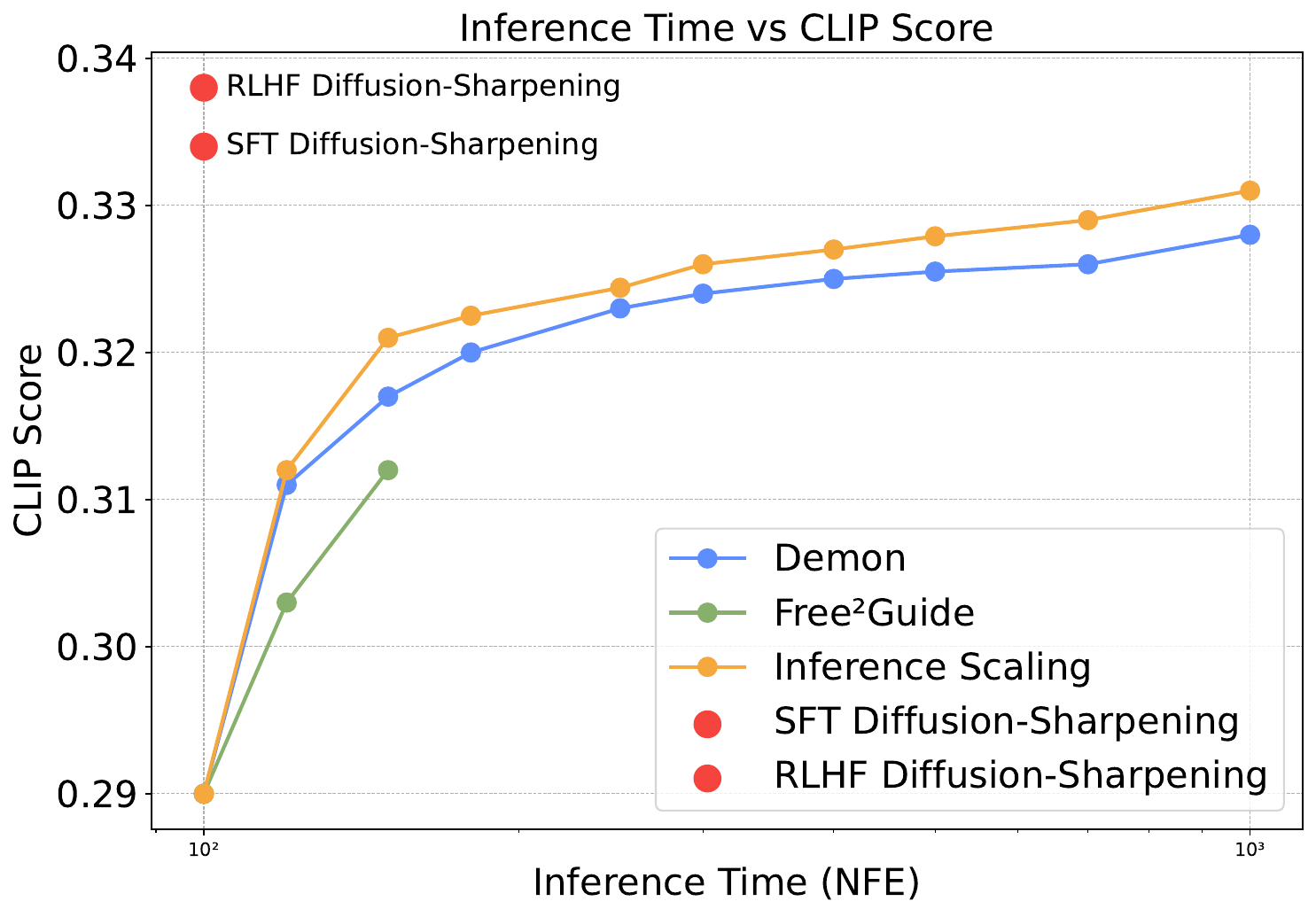}
    \vspace{-2em}
    \caption{Inference Performance of Diffusion Sharpening.}
    \label{fig:infer_cost}
    \vspace{-0.8em}
\end{figure}
Beyond training efficiency, our method also achieves optimal inference performance. Using CLIP Score as the reward model during inference, we evaluate SDXL with the default 100 NFE. 
As shown in \cref{fig:infer_cost}, all sampling trajectory optimization methods improve performance as NFE increases. However, the computational cost rises sharply, with methods like Demon and Inference Scaling requiring over 10,000 NFE, leading to inference times of several hours per image—rendering them impractical for real-world use. In contrast, our method integrates inference optimization into training, focusing on refining the sampling trajectory. This allows it to achieve superior performance within the same inference time as the baseline SDXL, demonstrating its efficiency.

\subsection{Ablation Study}
\begin{figure}[ht]
    \centering
    \includegraphics[width=\linewidth]{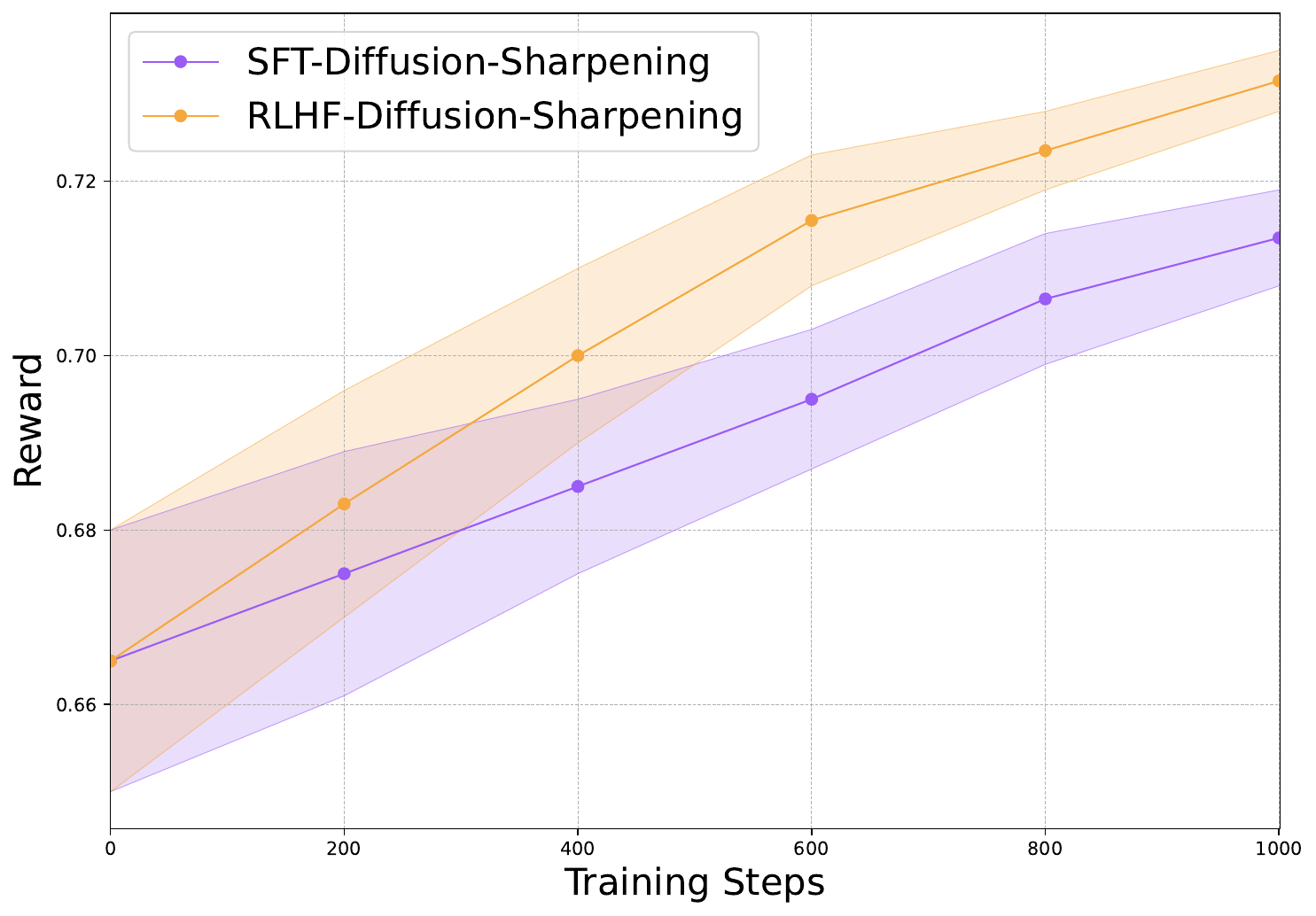}
    \vspace{-2em}
    \caption{Diffusion Sharpening Fine-tuning Reward Curve.}
    \label{fig:trajectory}
    \vspace{-1.3em}
\end{figure}
\paragraph{Effect of Sampling Trajectory Optimization}
To validate the optimization process along the sampling trajectory, we conducted an ablation study focusing on Sampling Trajectory Optimization. During training, we log reward results for both SFT and RLHF Diffusion Sharpening. As shown in \cref{fig:trajectory}, we track reward scores over the first 1000 SDXL fine-tuning steps, with the shaded region representing the standard deviation across multiple sampled trajectories at each step. The results show a steady increase in average reward and a decrease in variance as training progresses, indicating the model’s convergence toward more optimal paths. This confirms the effectiveness of our approach in enhancing both stability and performance during training.

\paragraph{Analysis of the Number of Samples}
We analyze the effect of the sampling number of samples $n$ during training and set the number of steps $m=1$ for comparison. As is shown in \cref{tab:traj_sample}, a number of samples of 1 corresponds to a standard DPO finetuning pipeline and we find an optimal number of samples = 3 for the final training configuration.
\begin{table}[ht]
    \centering
    \vspace{-1.5em}
    \caption{Performance of Different Number of Samples in Training}
    \resizebox{\linewidth}{!}{
    \begin{tabular}{cccc}
    \toprule
    \textbf{Number of Steps} & \textbf{CLIP Score}  & \textbf{ImageReward} & \textbf{MLLM} \\
    \hline
    1  & 0.334 & 1.352 & 0.864 \\
2  & 0.336 & 1.355 & 0.891\\
\textbf{3}  & \textbf{0.338} & 1.445 &\textbf{0.921} \\
4  & 0.336 & \textbf{1.446} &  0.911\\
8  & 0.337 & 1.444 & 0.919\\
\bottomrule
    \end{tabular}}
    \vspace{-1.5em}
    \label{tab:traj_sample}
\end{table}

\paragraph{Analysis of the Number of Steps}
We also analyze the effect of the sampling number of steps $m$ during training in \cref{tab:traj} after choosing the number of samples $n$.
A number of steps of 1 corresponds to a standard end-to-end fine-tuning baseline. The results show that increasing the number of steps leads to improved model performance. We set the number of steps to 3 for balancing cost and performance.
\begin{table}[ht]
    \centering
    \vspace{-1.5em}
    \caption{Performance of Different Number of Steps in Training}
    \resizebox{\linewidth}{!}{
    \begin{tabular}{cccc}
    \toprule
    \textbf{Number of Steps} & \textbf{CLIP Score}  & \textbf{ImageReward} & \textbf{MLLM} \\
    \hline
    1  & 0.322 & 1.321 & 0.897 \\
2  & 0.328 & 1.357 & 0.902\\
\textbf{3}  & \textbf{0.338} & \textbf{1.445} &0.921 \\
4  & 0.334 & 1.442 &  \textbf{0.923}\\
8  & 0.321 & 1.376 & 0.912\\
\bottomrule
    \end{tabular}}
    \vspace{-1.5em}
    \label{tab:traj}
\end{table}

\section{Conclusion}
% \subsection{Ablation Study}
In this work, we propose \textbf{Diffusion-Sharpening}, a novel fine-tuning approach that optimizes diffusion model performance by refining sampling trajectories. Our method addresses the limitations of existing approaches by enabling trajectory-level optimization through alignment with arbitrary reward models, while effectively amortizing the high inference costs. We introduce two variants: \textbf{SFT-Diffusion-Sharpening}, which leverages supervised fine-tuning for efficient backward trajectory optimization, and \textbf{RLHF-Diffusion-Sharpening}, which eliminates the need for curated datasets and performs online trajectory optimization. Through extensive experiments,  we demonstrate superior training efficiency as well as inference efficiency. Across diverse metrics, our \textbf{Diffusion-Sharpening} consistently outperforms existing fine-tuning methods and sampling trajectory optimization approaches.


\bibliography{example_paper}
\bibliographystyle{icml2025}

%%%%%%%%%%%%%%%%%%%%%%%%%%%%%%%%%%%%%%%%%%%%%%%%%%%%%%%%%%%%%%%%%%%%%%%%%%%%%%%
%%%%%%%%%%%%%%%%%%%%%%%%%%%%%%%%%%%%%%%%%%%%%%%%%%%%%%%%%%%%%%%%%%%%%%%%%%%%%%%
% APPENDIX
%%%%%%%%%%%%%%%%%%%%%%%%%%%%%%%%%%%%%%%%%%%%%%%%%%%%%%%%%%%%%%%%%%%%%%%%%%%%%%%
%%%%%%%%%%%%%%%%%%%%%%%%%%%%%%%%%%%%%%%%%%%%%%%%%%%%%%%%%%%%%%%%%%%%%%%%%%%%%%%
\newpage
\appendix
\onecolumn

% You can have as much text here as you want. The main body must be at most $8$ pages long.
% For the final version, one more page can be added.
% If you want, you can use an appendix like this one.  

% The $\mathtt{\backslash onecolumn}$ command above can be kept in place if you prefer a one-column appendix, or can be removed if you prefer a two-column appendix.  Apart from this possible change, the style (font size, spacing, margins, page numbering, etc.) should be kept the same as the main body.
%%%%%%%%%%%%%%%%%%%%%%%%%%%%%%%%%%%%%%%%%%%%%%%%%%%%%%%%%%%%%%%%%%%%%%%%%%%%%%%
%%%%%%%%%%%%%%%%%%%%%%%%%%%%%%%%%%%%%%%%%%%%%%%%%%%%%%%%%%%%%%%%%%%%%%%%%%%%%%%
\section{Implemantation Details}
\subsection{Baseline Models Configuration}
\label{sec:baseline}
% \subsection{Baseline Models Configuration}
In this section, we describe the configurations of different baseline models used in our study. We adopt the original model implementations whenever possible. For models that are not open-sourced or not directly compatible with SDXL, we perform minimal adaptations based on the original papers.

\begin{itemize}
    \item \textbf{Diffusion-DPO~\citep{wallace2024diffusion}:}Re-formulate Direct Preference Optimization (DPO) for diffusion models by incorporating a likelihood-based objective, utilizing the evidence lower bound to derive a differentiable optimization process. Using the Pick-a-Pic dataset containing 851K crowdsourced pairwise preferences, they fine-tuned the base SDXL-1.0 model with Diffusion-DPO. We directly use the pretrained Diffusion-DPO on SDXL.
    
    \item \textbf{DDPO~\citep{black2024training}:} This method optimizes diffusion models directly on downstream objectives using reinforcement learning (RL). By framing denoising diffusion as a multi-step decision-making process, it enables policy gradient algorithms referred to as Denoising Diffusion Policy Optimization. We adapt the original implementation to SDXL and use aesthetic quality as the optimization metric for fine-tuning.

    \item \textbf{D3PO~\citep{yang2024using}:} This approach omits the training of a reward model and instead functions as an optimal reward model trained using human feedback data to guide the learning process. By eliminating the need for explicit reward model training, D3PO proves to be a more direct and computationally efficient solution.

    \item \textcolor{black}{\textbf{IterPO~\citep{zhang2024itercomp}:} This method is the alignment framework of IterComp~\citep{zhang2024itercomp}, which collects composition-aware model preferences from multiple models and employ an iterative feedback learning approach to enable the progressive self-refinement of both the base diffusion model and reward models.}

    \item \textbf{Demon~\citep{yeh2024training}:} This method guides the denoising process at inference time without backpropagation through reward functions or model retraining. We adapt the original method to SDXL using the EDM scheduler with the \texttt{tanh-demon} configuration, setting a fixed inference cost of five minutes per image generation.

    \item \textbf{Free$^{2}$Guide~\citep{kim2024free}:} A gradient-free framework for aligning generated videos with text prompts without requiring additional model training. Leveraging principles from path integral control, Free$^{2}$Guide approximates guidance for diffusion models using non-differentiable reward functions. Since the original work focuses on video models, we directly adapt the provided pseudo-code to SDXL with a DDIM scheduler. Experiments are conducted on randomly selected $T=5$ inference steps, maintaining the same reward model settings.

    \item \textbf{Inference-Scaling~\citep{ma2025inference}:} This method formulates a search problem to identify better noise initializations for the diffusion sampling process. The design space is structured along two axes: the verifiers providing feedback and the algorithms searching for optimal noise candidates. While the original paper evaluates this approach on FLUX.1-DEV, we adapt the pseudo-code to SDXL, maintaining a fixed inference cost of five minutes and using the same verifier configurations for evaluation.
\end{itemize}

% \subsection{Dataset}
% For the text-to-image fine-tuning task, we utilize the following datasets:
\subsection{Datasets}
\label{app:dataset}
We utilize multiple datasets for training and evaluation, covering a diverse range of text-to-image tasks. Below, we describe each dataset used in our experiments:
\begin{itemize}
    \item \textbf{JourneyDB}~\citep{pan2023journeydb}: A large-scale collection of high-resolution images generated by Midjourney. This dataset contains diverse and detailed text descriptions that capture a wide range of visual attributes, enabling robust multi-modal training.

    \item \textbf{Text-to-Image-2M~\citep{zk_2024}}: A curated text-image pair dataset designed for fine-tuning text-to-image models. The dataset consists of approximately 2 million samples, carefully selected and enhanced to meet the high demands of text-to-image model training.

    \item \textbf{Pokemon-Blip~\citep{pokeman}}: A dataset containing unique Pokémon images labeled with BLIP-generated captions. It is specifically designed to evaluate adaptation to seen data and assess the model’s convergence capabilities.

    \item \textbf{DiffusionDB}~\citep{wang2022diffusiondb}: The first large-scale text-to-image prompt dataset, containing 14 million images generated by Stable Diffusion using user-specified prompts and hyperparameters. The unprecedented scale and diversity of this human-actuated dataset provide valuable research opportunities in understanding the interplay between prompts and generative models, detecting deepfakes, and designing human-AI interaction tools to improve model usability.

    \item \textbf{DrawBench}~\citep{saharia2022photorealistic}: A comprehensive and challenging benchmark for text-to-image models, introduced by the Imagen research team. It consists of 200 prompts spanning 11 diverse categories. The benchmark evaluates text-to-image models’ ability to handle complex prompts and generate realistic, high-quality images. During evaluation, we generate one image per prompt.
\end{itemize}

\subsection{Training Settings}
We train our models with carefully optimized settings to ensure stable and efficient training. We use the AdamW optimizer without weight decay, configured with beta parameters \((\beta_1=0.0, \beta_2=0.99)\). The learning rate is set to \(5 \times 10^{-6}\), reflecting the distinct requirements of each modality. Both diffusion sharpening models are trained with a batch size of 8.

\subsection{Evaluation Settings}
For \textit{MLLM Grader}, we prompt the GPT-4o model to assess synthesized images from five different perspectives: Accuracy to Prompt, Originality, Visual Quality, Internal Consistency, and Emotional Resonance following ~\citep{ma2025inference}. Each perspective is rated from $0$ to $100$, and the averaged overall score is used as the final metric. In Figure~\ref{fig:gemini-prompt} we present the detailed prompt. We observe that search can be beneficial to each scoring category of the \textit{MLLM Grader}.

\textbf{T2I-CompBench.} For each prompt we search for two noises and generate two samples. During evaluation, the samples are splitted into six categories: \texttt{color},  \texttt{shape}, \texttt{texture}, \texttt{spatial}, \texttt{numeracy}, and \texttt{complex}. Following~\citet{huang2023t2i}, we use the BLIP-VQA model for evaluation in \texttt{color},  \texttt{shape}, and \texttt{texture}, the UniDet model for \texttt{spatial} and \texttt{numeracy}, and a weighted averaged scores from BLIP VQA, UniDet, and CLIP for evaluating the \texttt{complex} category.

% \subsection{Hyperparameter Settings}

\section{User Study}
\label{app:user}
\begin{figure}[ht]
    \centering
    \includegraphics[width=\linewidth]{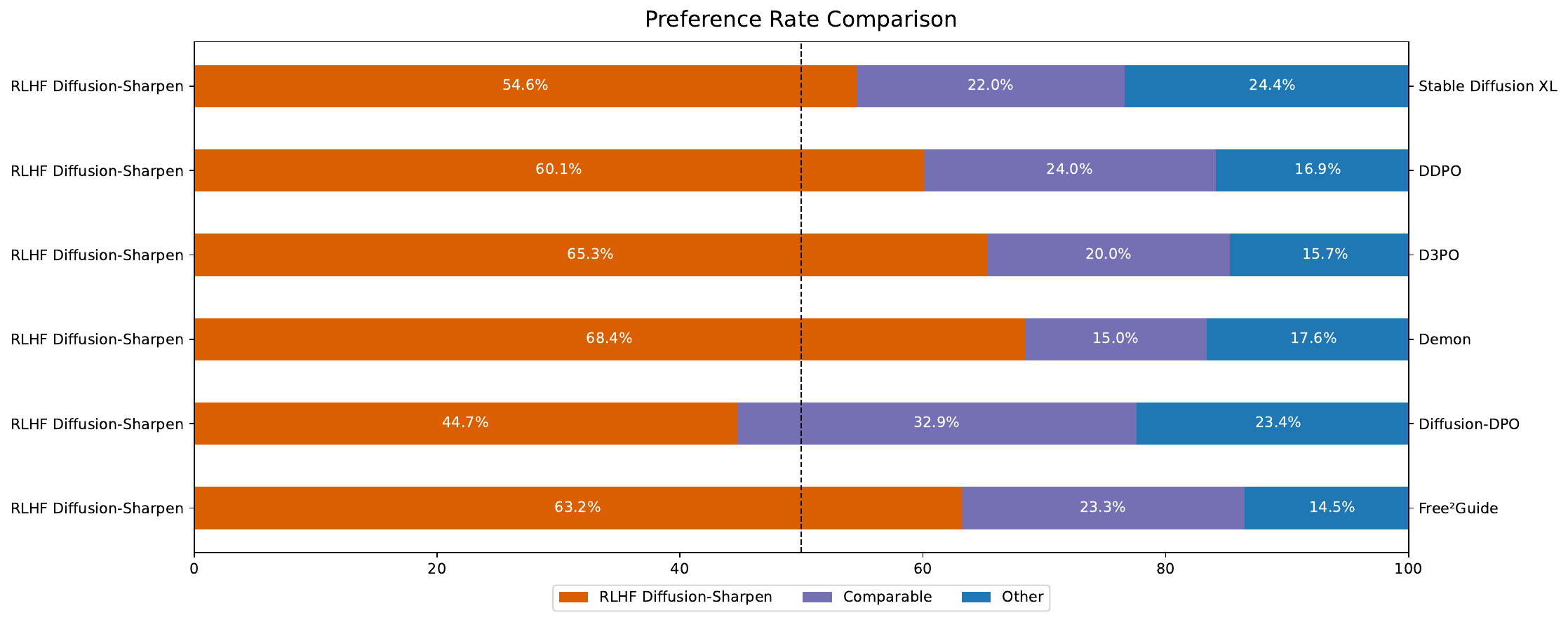}
    % \vspace{-2em}
    \caption{User Study about Comparision with Other Methods}
    \label{fig:user}
    % \vspace{-2em}
\end{figure}

To verify the effectiveness of our proposed Diffusion-Sharpening, we conduct an extensive user study across various scenes and models. Users compared model pairs by selecting their preferred video from three options: method 1, method 2, and comparable results. 
As presented in \cref{fig:user}, our method (orange in left) obtains more user preferences than others (blue in right), which further proving its effectiveness.

\clearpage

\section{MLLM Grader Design}
\label{app:mllm}
\begin{figure}[htb!]
    \centering
    \fbox{
        \begin{minipage}{\textwidth}
            \ttfamily
            "You are a multimodal large-language model tasked with evaluating images generated by a text-to-image model. Your goal is to assess each generated image based on specific aspects and provide a detailed critique, along with a scoring system. The final output should be formatted as a JSON object containing individual scores for each aspect and an overall score. Below is a comprehensive guide to follow in your evaluation process:

            \textbf{1. Key Evaluation Aspects and Scoring Criteria:}
            
            For each aspect, provide a score from 0 to 10, where 0 represents poor performance and 10 represents excellent performance. For each score, include a short explanation or justification (1-2 sentences) explaining why that score was given. The aspects to evaluate are as follows:

            \textbf{a) Accuracy to Prompt}

            Assess how well the image matches the description given in the prompt. Consider whether all requested elements are present and if the scene, objects, and setting align accurately with the text.
            Score: 0 (no alignment) to 10 (perfect match to prompt).

            \textbf{b) Creativity and Originality}

            Evaluate the uniqueness and creativity of the generated image. Does the model present an imaginative or aesthetically engaging interpretation of the prompt? Is there any evidence of creativity beyond a literal interpretation?
            Score: 0 (lacks creativity) to 10 (highly creative and original).

            \textbf{c) Visual Quality and Realism}
            
            Assess the overall visual quality, including resolution, detail, and realism. Look for coherence in lighting, shading, and perspective. Even if the image is stylized or abstract, judge whether the visual elements are well-rendered and visually appealing.
            Score: 0 (poor quality) to 10 (high-quality and realistic).

            \textbf{d) Consistency and Cohesion}

            Check for internal consistency within the image. Are all elements cohesive and aligned with the prompt? For instance, does the perspective make sense, and do objects fit naturally within the scene without visual anomalies?
            Score: 0 (inconsistent) to 10 (fully cohesive and consistent).

            \textbf{e) Emotional or Thematic Resonance}

            Evaluate how well the image evokes the intended emotional or thematic tone of the prompt. For example, if the prompt is meant to be serene, does the image convey calmness? If it’s adventurous, does it evoke excitement?
            Score: 0 (no resonance) to 10 (strong resonance with the prompt's theme).
            
            \textbf{2. Overall Score}

            After scoring each aspect individually, provide an overall score, representing the model's general performance on this image. This should be a weighted average based on the importance of each aspect to the prompt or an average of all aspects."
        \end{minipage}
    }
    \caption{\footnotesize{\textbf{\textit{The detailed prompt for evaluation with the MMLLM Grader.}}}}
    \label{fig:gemini-prompt}
\end{figure}

\clearpage
\section{More Qualitative Results}

% \begin{figure}[ht]
%     % \setlength{\abovecaptionskip}{0pt}
%     % \setlength{\belowcaptionskip}{-3pt}
%     \centering
%     \includegraphics[width=\linewidth]{figs/fig1-1.pdf}
%     \vspace{-2em}
%     \caption{Comparison of Three Diffusion-Based Methods for Reward-Driven Optimization: (i) Diffusion Reinforcement Learning, (ii) Diffusion Sampling Trajectory Optimization, and (iii) Diffusion Sharpening.}
%     \label{fig:intro}
%     \vspace{-2em}
% \end{figure}

\begin{figure}[htb!]
    \centering
    \includegraphics[width=0.61\linewidth]{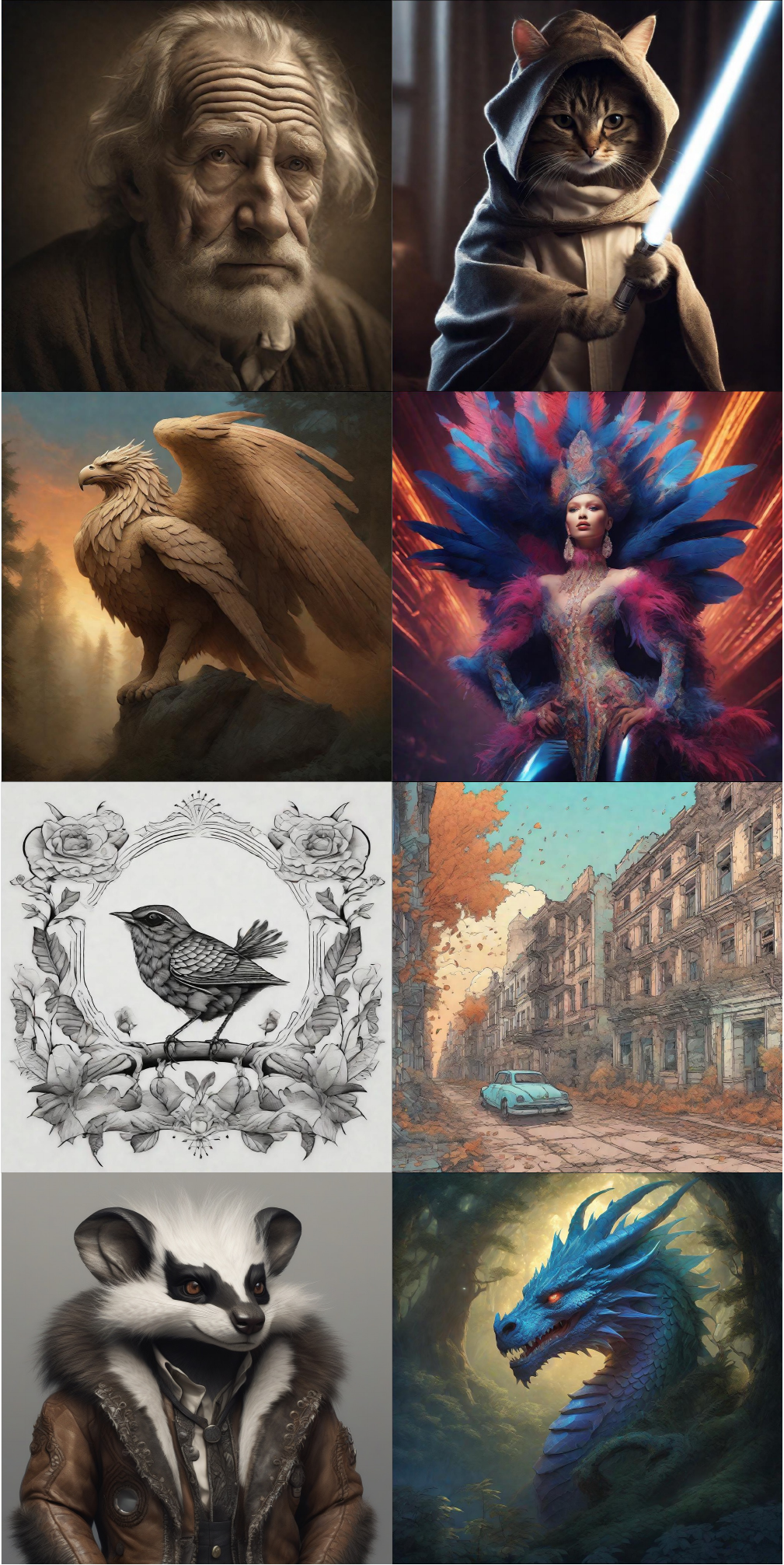}
    % \vspace{-2em}
    \caption{More Qualitative Results for SFT Diffusion-Sharpening.}
    \label{fig:app_more_results}
\end{figure}

\begin{figure}[htb!]
    \centering
    \includegraphics[width=0.9\linewidth]{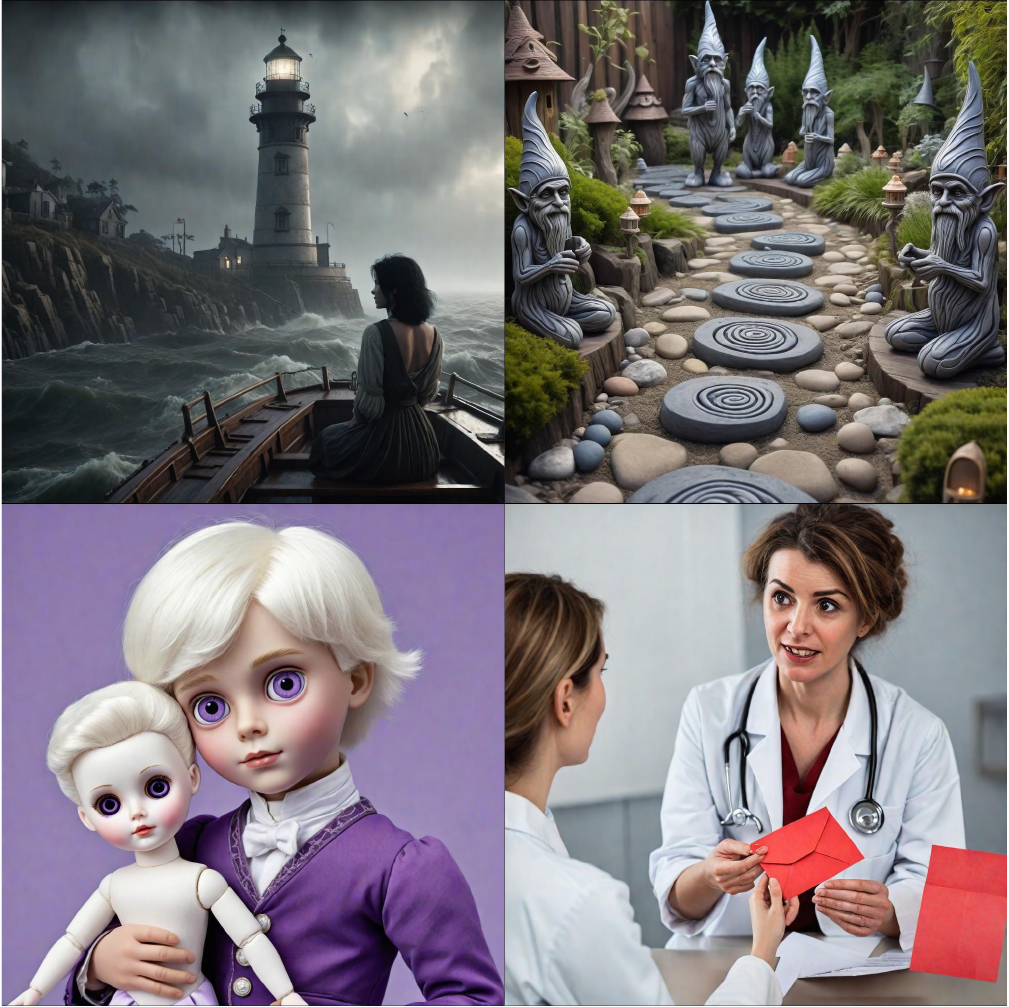}
    % \vspace{-2em}
    \caption{More Qualitative Results for RLHF Diffusion-Sharpening.}
    \label{fig:app_more_results2}
\end{figure}
\end{document}